\documentclass[conference]{IEEEtran}
\IEEEoverridecommandlockouts
\usepackage{cite}
\usepackage{amsmath,amssymb,amsfonts}
\usepackage{algorithm}
\usepackage{algorithmic}
\usepackage{graphicx}
\usepackage{textcomp}
\usepackage[T1]{fontenc}
\usepackage{xcolor}
\def\BibTeX{{\rm B\kern-.05em{\sc i\kern-.025em b}\kern-.08em
    T\kern-.1667em\lower.7ex\hbox{E}\kern-.125emX}}
\begin{document}

\title{Learning from Event Cameras with Sparse\\ Spiking Convolutional Neural Networks 
\thanks{This material is based upon work supported by the French ANRT through a CIFRE PhD thesis in collaboration with Renault.}
}

\author{\IEEEauthorblockN{Loïc Cordone}
\IEEEauthorblockA{\textit{Renault and LEAT / CNRS UMR 7248} \\
Sophia Antipolis, France \\
loic.cordone@renault.com}
\and
\IEEEauthorblockN{Benoît Miramond}
\IEEEauthorblockA{\textit{LEAT / CNRS UMR 7248} \\
\textit{University Côte d'Azur}\\
Sophia Antipolis, France \\
benoit.miramond@univ-cotedazur.fr}
\and
\IEEEauthorblockN{Sonia Ferrante}
\IEEEauthorblockA{\textit{Renault}\\
Sophia Antipolis, France \\
sonia.ferrante@renault.com}
}

\maketitle

\begin{abstract} Convolutional neural networks (CNNs) are now the de facto solution for computer vision problems thanks to their impressive results and ease of learning. These networks are composed of layers of connected units called artificial neurons, loosely modeling the neurons in a biological brain. However, their implementation on conventional hardware (CPU/GPU) results in high power consumption, making their integration on embedded systems difficult. In a car for example, embedded algorithms have very high constraints in term of energy, latency and accuracy. To design more efficient computer vision algorithms, we propose to follow an end-to-end biologically inspired approach using event cameras and spiking neural networks (SNNs). Event cameras output asynchronous and sparse events, providing an incredibly efficient data source, but processing these events with synchronous and dense algorithms such as CNNs does not yield any significant benefits. To address this limitation, we use spiking neural networks (SNNs), which are more biologically realistic neural networks where units communicate using discrete spikes. Due to the nature of their operations, they are hardware friendly and energy-efficient, but training them still remains a challenge. Our method enables the training of sparse spiking convolutional neural networks directly on event data, using the popular deep learning framework PyTorch. The performances in terms of accuracy, sparsity and training time on the popular DVS128 Gesture Dataset make it possible to use this bio-inspired approach for the future embedding of real-time applications on low-power neuromorphic hardware.
\end{abstract}

\section{Introduction}
Event cameras are bio-inspired sensors composed of independent photoreceptor pixels that emit binary events when they detect a brightness change. Unlike traditional cameras that output a sequence of frames, these cameras output an asynchronous stream  of events. This principle of operation provides several advantages: a high temporal resolution (microseconds) and a high dynamic range (>120dB) while producing up to 1000 times fewer data and consuming less than 100mW. These cameras are therefore perfectly suited for embedded use, as long as we can develop applications that take advantage of their unconventional output. Indeed, frame-based vision algorithms designed for image sequences are not directly applicable to event data. 

In order to take advantage of the very low latency of event cameras in an embedded real-time application, it is necessary to use an algorithm capable of directly processing the flow of binary events. Spiking neural networks, because of the nature of their processing, represent a promising lead. Training large spiking neural networks is an active area of research, with the majority of these works looking for the highest possible accuracy, often comparing it to that of classical neural networks. Although accuracy is an important metric when evaluating a solution, it is not the only one, especially when working on real-time applications. In this work, we present a different approach to the training of spiking neural networks on event data, by focusing on metrics essential to their embeddeding on specialized low-power hardware, like the sparsity of the inference or the memory footprint of the network (number of parameters)

The main contributions of this paper can be summarized as follows:
\begin{enumerate}
    \item We propose an improvement over the supervised backpropagation-based learning algorithm using spiking neurons presented in \cite{neftci} and \cite{s2net} with the use of strided sparse convolutions. The training implementation is also modified to allow timestep-wise rather than layer-wise learning. This greatly reduces the training time while generating sparser networks during inference, with higher accuracy.
    \item We investigate a full event-based approach composed of sparse spiking convolutions, respecting the data temporality across the network and the constraints of an implementation on specialized neuromorphic and low power hardware.
    \item We evaluate our approach on the neuromorphic DVS128 Gesture dataset \cite{dvs128}, achieving competitive results while using a much smaller and sparser network than other spiking neural networks.
\end{enumerate}

Our code is available upon request and will be available online in the future.

\section{Related work}

\subsection{Learning from event sensors}

Processing event-by-event data coming directly from an event sensor is typically addressed by two kinds of methods: filters (determinitisc or probabilistic) and multi-layer artificial neural networks. While the former is interesting for its simplicity, the latter represents the majority of state-of-the-art results on event data. 

The simplest way to use event data for supervised learning is to accumulate events pixel-wise over a period of time, either by counting them or by accumulating their polarities, essentially creating an event frame, also called a 2D histogram \cite{steering2018}. Other approaches include: 2D time surfaces \cite{hots}, where each pixel stores the timestamp of its last event ; motion-compensated event images, where events are accumulated on a frame and sharpened using a motion-compensation realignement \cite{iwe} ; and reconstructed images, where an artificial neural network reconstructs images directly from events \cite{reconstruction}. While all these methods allow the use of traditional computer vision algorithms, they need a certain amount of preprocessing and they change the nature of the data, especially its temporality. Voxel grids are another popular event data representation used for supervised learning with CNNs, in which each voxel represents a pixel and a time interval \cite{voxel}. This representation better preserves temporal information but requires more memory and more computations. It is also the preferred representation for spiking neural networks, because the events can be interpreted as spikes emitted at a certain time, suppressing the need for any preprocessing. Voxels grids with different accumulation policies are used in \cite{slayer}, \cite{scrnn}, \cite{decolle} and \cite{fangtim}.

\subsection{Spiking neural networks}

Since their introduction a few decades ago \cite{snn}, spiking neural networks have generated a lot of interest from both the neuroscience and the artificial intelligence fields. The architecture of an SNN consists of spiking neurons connected by adjustable scalar weights, modeling the synapses. There are many different models of a spiking neuron, from the simplest, the Leaky Integrate-and-Fire (LIF) \cite{lif}, to the more biologically realistic, the Hodgkin-Huxley neuron \cite{hh}. Due to its simplicity, the LIF neuron is widely used, and also because using more realistic neurons would require more computing power that is not compatible with the low-power appeal of SNNs. 

As it is the case for traditional neural networks, it is possible to learn the scalar weights. However, spikes are discrete and thus non differentiable, which prevents the use of the popular back-propagation algorithm on SNNs. As a result, multiple learning rules have been proposed. The most bio-plausible one is spike-timing-dependent plasticity (STDP), an unsupervised learning rule where the weight connecting two neurons is modified according to the delay between the firing of the presynaptic and the postsynaptic neurons. It is common to use STDP for the unsupervised learning of an SNN and to use a supervised classifier (SVM, neural networks) to obtain the final predictions \cite{kherad}. Although results are improving rapidly, STDP is not competitive with supervised learning for datasets bigger than MNIST. 

Several supervised learning rules have been proposed to train SNNs, among them we can mention SpikeProp, SLAYER \cite{slayer} and the ones using a surrogate gradient \cite{neftci}, \cite{s2net}. SpikeProp \cite{spikeprop} was one of the first algorithm to train SNNs by using a sort of backpropagation. They used a particular model for the spiking neurons where the outputs can be modeled as continuous values, enabling the calculation of derivatives. The errors were backpropagated based on membrane potentials at spike times. However, with this work each neuron can only output one spike. SLAYER \cite{slayer} introduced a new backpropagation algorithm for learning synaptic weights and axonal delays, with excellent results on neuromorphic datasets. 

More recently, \cite{neftci} introduced the concept of surrogate gradient for the learning in SNNs. In this approach, spikes are generated using a particular activation function, which is a simple Heaviside step function during the forward pass, but in the backward pass the gradient is determined using a surrogate gradient, for example the gradient of a sigmoid function. This method relies on the fact that the surrogate function approximates the Heaviside function, thus its derivative provides a sufficient approximation for learning with backpropagation. This work also proves the equivalence between RNNs and SNNs, allowing their learning in popular DNNs frameworks using BackPropagation Through Time (BPTT). Reference \cite{s2net} extends the mechanisms proposed by \cite{neftci} by enabling the learning of spiking convolutional networks. Our work is mainly based on these two methods.

\subsection{Sparse convolutional networks}

\begin{figure}[t]
\centering
{\includegraphics[width=1.5in]{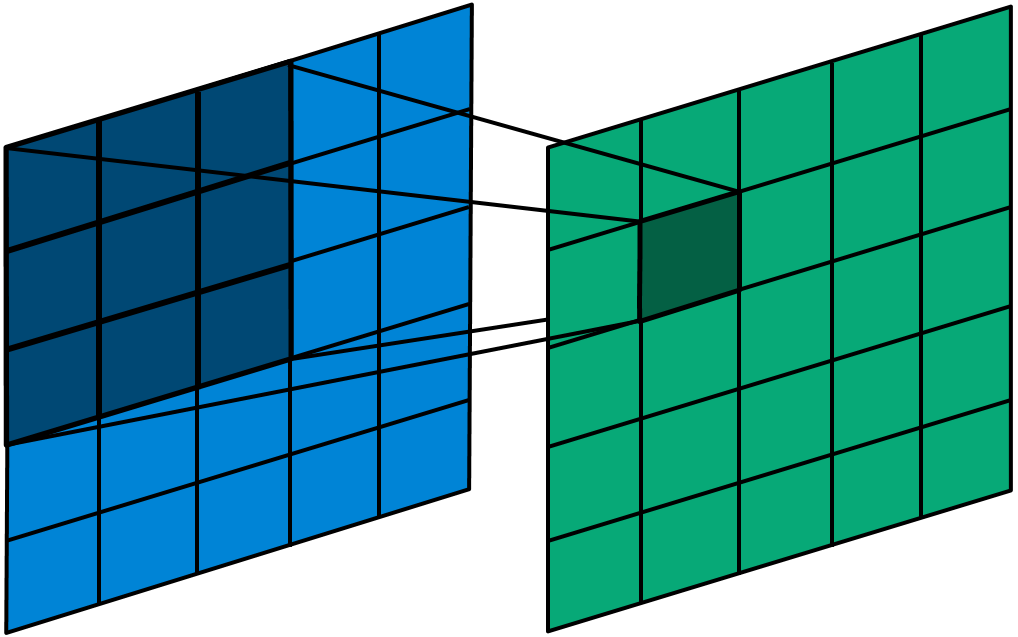}%
\label{fig_denseconv}}
\hfil
{\includegraphics[width=1.5in]{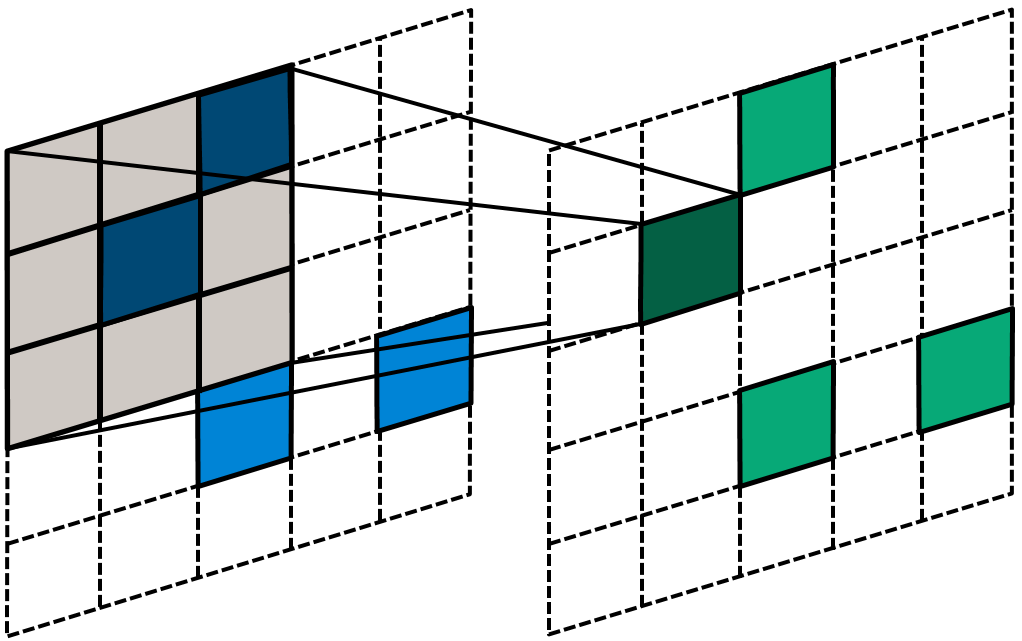}%
\label{fig_sparseconv}}
\caption{Vizualisation of a convolution on a dense tensor and on a sparse tensor. The number of operations and the results differ. More details can be found in \cite{me}.}
\label{fig_densesparseconv}
\end{figure}

A well studied method to speedup inference and minimize memory footprint is to prune the weights in CNNs \cite{sparseCNN}, inducing sparsity in the model parameters. However, input data and activations remain dense. It also exists sparse convolutional networks that conserve the parameters dense but are able to process spatially sparse data, generating sparse tensors throughout the network. The immediate benefit of this approach is that the processing is done only on non-zero data, unlike dense CNNs, which is particularly interesting for highly sparse data such as the output of an event camera (e.g. the DVS Gesture dataset samples we used were 99\% sparse). Another benefit is that sparse convolutions maintain the sparsity of the input data across layers, resulting in a sparse network (see Fig.~\ref{fig_densesparseconv}).

The main libraries to train sparse convolutional networks are SparseConvNet \cite{sparseconvnet} and Minkowski Engine \cite{me}, both are GPU-accelerated and based on PyTorch. Sparse CNNs have been used primarily on LIDAR and depth cameras data \cite{me}, but most recent works include their use with event data. Reference \cite{eventscnn} trained CNNs with SparseConvNet on N-Caltech101 and N-Cars with similar accuracy as the dense network with up to 20 times less computations per event. To the best of our knowledge, this paper presents the first work to use a sparse spiking convolutional network to learn directly from spatio-temporal event data.

\section{Method}

\subsection{Event-based data}
The most popular event cameras have a temporal resolution of 1 microsecond and a spatial resolution that can vary from $128\times128$ to $1280\times720$ pixels \cite{dvs128camera}, \cite{prophesee1Megapixel}. To our knowledge, no satisfactory method allows to use the data produced by this type of sensor for machine learning while maintaining a temporal resolution of 1 microsecond. We therefore accumulate the events on larger time windows $\Delta t$ seconds, while keeping the spatial resolution unchanged, essentially constituting a voxel grid. For a recording of events of duration $d$ seconds, we have thus gathered the events on $\frac{d}{\Delta t} = T$ timesteps. The event data are stored in the form of a 4D $CTHW$ tensor, with $C$ the number of channels, $T$ the number of timesteps, $H$ and $W$ the height and width of the data. Using this data representation, we can build conventional 3D convolutional networks, where the time dimension is considered as a spatial dimension, or 2D spiking convolutional networks operating on each timestep sequentially.

It is important to note that unlike other works, our voxel grid retains the events temporal information and that the accumulation on time windows is binary: we do not sum the events nor save their precise timestamps, we only record if at least one event has been emitted in the time window $\Delta t$. This constraint leads to a loss of information but it is the only one compatible with a real-time use case where the system would process the flow of events on the fly. With this choice, it is as if we had modified the temporal resolution of the event camera to be $\Delta t$.

\subsection{Models architecture}

Our models are constructed by stacking convolutional layers with a growing number of filters, as it is common to do in machine learning. We reduce the spatial dimensions of feature maps through the network with strided convolutions. We used a stride of 2 rather than using pooling layers because implementing the pooling operation in the spike domain on specialized hardware remains difficult, even if recent works show promising leads \cite{pool1}, \cite{pool2}. The comparison and the evaluation of strided convolutions over pooling is made in the section \ref{stride}. As for the time dimension, it is not reduced through the network but kept constant. This constraint allows the network to produce output feature maps for each single timestep. 

The output feature maps are then flattened along the filter and spatial dimensions, resulting in a $T \times s$ 2D tensor, with $s$ being the product of the number of filters of the last convolution, the height and the width of the feature map. A final fully-connected layer then outputs $T$ predictions, one per timestep. The final prediction of the network can be obtained by any temporal reduction across the $T$ predictions. For training, we applied a simple mean as it gives the best accuracy. Other options would be to sum the predictions, or output a prediction once a class score has exceeded a threshold value. Our method thus does not constrain the number of timesteps needed for inference, as the network is able to output a final decision at any timestep if required. 

\subsection{Timestep-wise model}

This section will present the Leaky Integrate-and-Fire model and how we use it to do a timestep-wise learning of our SNNs.

An SNN is by definition a neural network composed of spiking neurons. We used Leaky Integrate-and-Fire (LIF) neurons as they provide a simple model of the behavior of the neuron membrane potential through time. The subthreshold dynamics of one LIF neuron can be defined as:

\begin{equation}
\label{eqn:lif}
    \tau_{mem} \frac{\mathrm{d}V(t)}{\mathrm{d}t} = -(V(t) - V_{rest}) + I(t),
\end{equation}

where $V(t)$ represents the neuron membrane potential at time $t$, $V_{rest}$ represents the resting potential, $\tau_{mem}$ represents the membrane time constant and $S(t)$ represents the input current to the neuron at time $t$. A spike is emitted when the membrane potential $V(t)$ reaches a certain threshold $V_{th}$. After each spike, the potential $V(t)$ is reset to the resting potential $V_{rest}$. 

Following the work done in \cite{s2net}, it is possible to approximate ~\eqref{eqn:lif} by linear recurrent equations in discrete time:

\begin{equation}
\label{eqn:rnn}
\begin{aligned}
    V_{rest}[n-1] &= V_{th}\Vert W\Vert^2 S[n-1], \\
    I[n] &= (1-\beta)S[n]\\
    V[n] &= \beta (V[n-1] - V_{rest}[n-1]) + I[n], \\
    S[n] &= \Theta(V[n] - V_{th}\Vert W\Vert^2)
\end{aligned}
\end{equation}

where $S(t)$ represents the input spikes to the neuron at time $t$, $\beta = \exp(-\frac{\Delta t}{\tau_{mem}})$ with $\Delta t$ the simulation time step, $\Theta$ is the Heaviside step function and $W$ is the synaptic weight matrix. Reference \cite{neftci} proved that~\eqref{eqn:rnn}, applied to neurons organized in layers, characterizes the dynamics of a RNN, as the computations used to update the cell state can be unrolled in time. Following \cite{s2net}, we considered $\beta$ and $b=V_{th}$ as trainable parameters which gives:
\begin{equation}
\label{eqn:real}
\begin{aligned}
    V[n] &= \beta (V[n-1] - V_{rest}[n-1]) + (1-\beta) S[n], \\
    S[n] &= \Theta(\frac{V[n]}{\Vert W\Vert^2 + \epsilon} - b),
\end{aligned}
\end{equation}

with $\epsilon = 1e^{-8}$. The potential is normalized by $\Vert W\Vert^2$ to avoid vanishing gradients during training. 

The gradient of the Heaviside step function is approximated by the gradient of a sigmoid function scaled by a parameter $\alpha \geq 0$:

\begin{equation}
\begin{aligned}
    \text{sig}_{\alpha} &= \frac{1}{1+exp(-\alpha x)}, \\
    \Theta'(x) &\approx \text{sig}_{\alpha}'(x) = \alpha \text{ sig}_{\alpha}(x) \text{ sig}_{\alpha}(-x)
\end{aligned}
\end{equation}

By organizing these spiking neurons in a two dimensional grid, we can construct spiking convolutional layers. Our spiking neural network algorithm is presented in Algorithm ~\ref{alg:snn}.

\begin{algorithm}[t]
\caption{Timestep-wise spiking neural network algorithm}
\begin{algorithmic}[1]
\label{alg:snn}
\renewcommand{\algorithmicrequire}{\textbf{Inputs:}}
\renewcommand{\algorithmicensure}{\textbf{Outputs:}}
\REQUIRE number of timesteps $T$, input event data $X = \{X_0,...,X_{T-1}\}$, number of convolutional layers $L$, spiking convolutional layers \{\texttt{sc\textsubscript{0},...,sc\textsubscript{L-1}}\}, dropout layer \texttt{d}, linear layer \texttt{fc} \\
\ENSURE  prediction output $Y$
\STATE create layer potentials $\{mem_0,...,mem_{L-1}\}$ initalized to 0, create an empty list $outs = \{\}$ 
\FOR {$t = 0$ to $T-1$}
\STATE input $X_t, mem_0$ to \texttt{sc\textsubscript{0}}, get $out_t, mem_0$
\FOR {$i = 1$ to $L-1$}
\STATE input $out_t, mem_i$ to \texttt{sc\textsubscript{i}}, get $out_t, mem_i$
\ENDFOR
\STATE input $out_t$ to \texttt{d}, get $out_t$
\STATE input $out_t$ to \texttt{fc}, get  $out_t$
\STATE append $out$ to $outs$
\ENDFOR
\RETURN $Y$ the mean of $outs$
\end{algorithmic}
\end{algorithm}

 The main difference with \cite{s2net} comes from our timestep-wise approach: here the potentials are initialized and updated at the model level rather than inside each layer. This makes the traversal of the network possible timestep per timestep, and reduces the training time by a factor of $L$, the number of convolutional layers. Our method is therefore totally independent of the number of timesteps, since it only processes one timestep at a time. With this algorithm and a suitable hardware, it would even be possible to process the output of an event camera without any change in its temporal resolution, i.e. every microsecond.

\subsection{Sparsity}

We chose to use sparse convolutions in our method because event data is highly sparse (>99\% sparse), thus processing only non-zero data promises to save time, memory and potentially to achieve a better learning, as already outlined by \cite{eventscnn}. Sparse convolutions have the advantage of keeping the sparsity of the data over the network, or at least of not damaging it as conventional convolutions would do. We measure this metric by counting the number of non-zero activations (spikes) outputted by each layer, and the percent of sparsity by dividing the number of non-zero activations by the output feature map size. Results are presented in Section \ref{results}.

\section{Experiments}

\subsection{Dataset}

\begin{figure}[t]
\centering
{\includegraphics[width=1.4in]{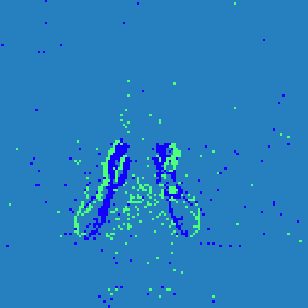}
\label{fig_handclap}}
\hfil
{\includegraphics[width=1.4in]{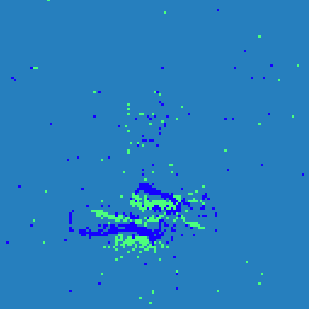}%
\label{fig_armroll}}
\caption{Examples of gestures from the DVS128 Gesture Dataset, where green and dark blue pixels correspond respectively to events that appear or disappear. This frame representation is obtained by the accumulation of events over a period of 20ms. The gestures depicted are a hand clap and an arm roll.}
\label{fig_dvsgesture}
\end{figure}

We evaluated our method on the IBM DVS128 Gesture Dataset \cite{dvs128}, a dataset published in 2017 which contains recordings of 29 subjects performing 11 hand gestures under 3 different illuminations. The gestures were recorded using a DVS camera with a resolution of $128\times128$ pixels (see Fig.~\ref{fig_dvsgesture}). Each gesture lasts up to 6 seconds, and are provided as raw events with a microsecond timestamp, an $x,y$ position and a polarity. Samples from the first 23 subjects are used for training and samples from the last 6 subjects for testing. 

Following \cite{slayer}, we provided only the first $1.5$s of each gesture to our network for both training and testing, with raw events accumulated over $T=150$ windows of $\Delta t = 10$ ms. These samples are composed of one input channel, the polarity, which is stored as either $+1$ or $-1$. The resulting voxel grid of shape $1\times150\times128\times128$ ($CTHW$) is stored as a Minkowski Engine sparse tensor for maximal efficiency. Apart from this change in representation, no pre-processing is applied to the events. \\

See Table \ref{tab:timesteps} for a comparison of the temporal characteristics of our model with other SNNs from the litterature. Our method offers a good compromise between the duration of computations (less than 150 timesteps) and the respect of the data temporality (timestep of 10ms). Reference \cite{fangtim} used a number of events to construct timesteps, which is not compatible with data coming from an event camera in real-time, as well as prevents from taking advantage of the data sparsity.

\begin{table}[b]
\renewcommand{\arraystretch}{1.3}
\centering
\caption{Comparison of the temporal characteristics with other SNNs}
\begin{tabular}{c|ccc}
\hline
\textbf{Model} & \multicolumn{1}{c}{\begin{tabular}[c]{@{}c@{}}\textbf{Timesteps}\\ \textbf{train/test}\end{tabular}} & \multicolumn{1}{c}{\begin{tabular}[c]{@{}c@{}} \textbf{Timestep} \\ \textbf{duration}\end{tabular}} &  \multicolumn{1}{c}{\begin{tabular}[c]{@{}c@{}}\textbf{Sample duration}\\ \textbf{train/test}\end{tabular}} \\ \hline
SLAYER \cite{slayer}  & 300 / 300 & 5ms & 1.5s / 1.5s \\  
SCRNN \cite{scrnn}    & 20 / 20 & 50ms & 1.0s / 1.0s \\ 
DECOLLE \cite{decolle} & 500 / 1800  & 1ms & 0.5s / 1.8s \\ 
PLIF SNN \cite{fangtim} & 20 / 20 & \multicolumn{1}{c}{\begin{tabular}[c]{@{}c@{}}N/A\\ (10k events/ts)\end{tabular}}  & \multicolumn{1}{c}{\begin{tabular}[c]{@{}c@{}}N/A\\ (400k events)\end{tabular}}\\
This work  & 150 / \textless{150} & 10ms & 1.5s / \textless{1.5s} \\ \hline
\end{tabular}
\label{tab:timesteps}
\end{table}

\subsection{Models}

\begin{figure*}[t]
\centering
\includegraphics[width=\textwidth]{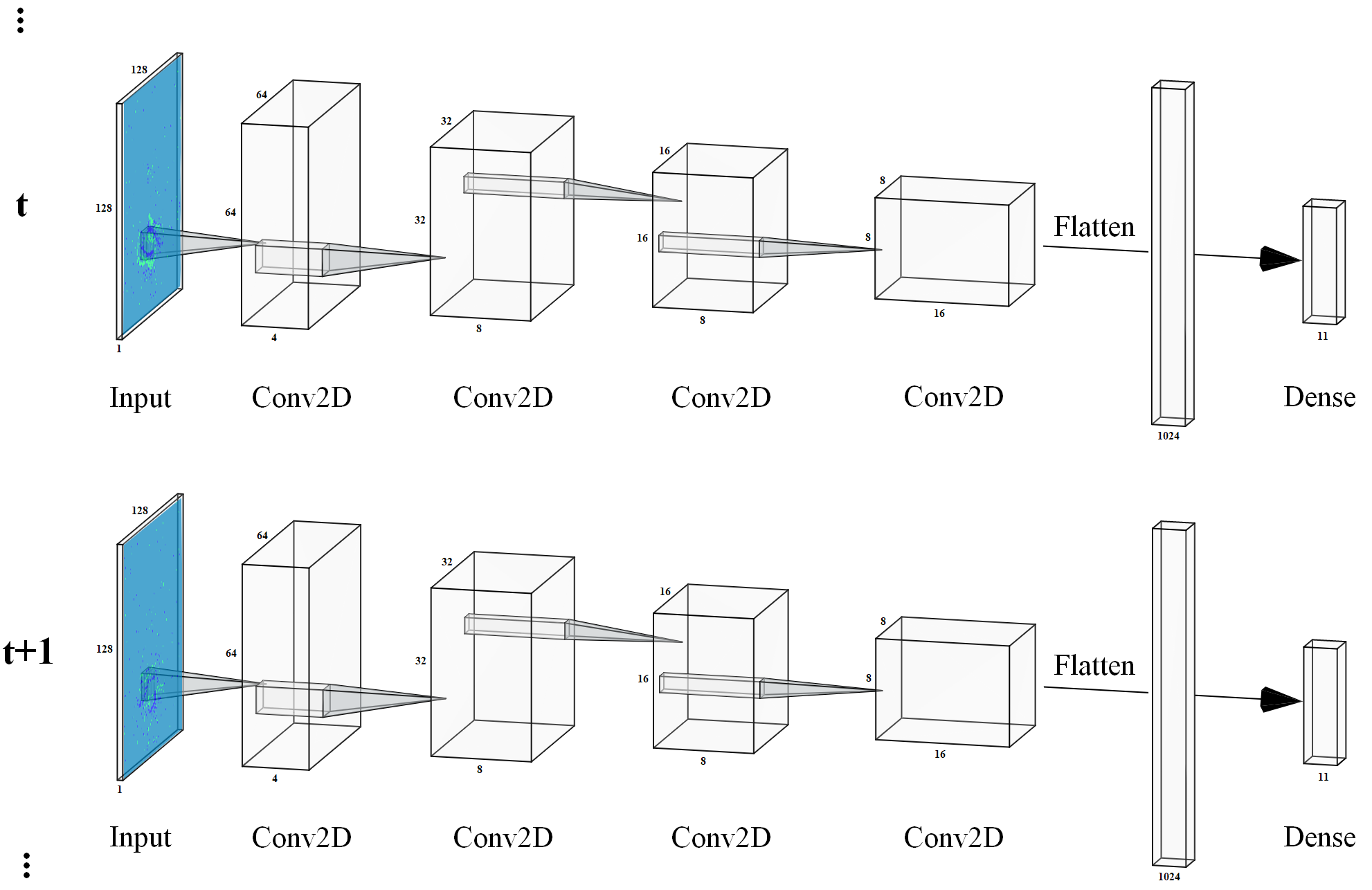}
\caption{SNN architecture. Each timestep of the input event data goes through the network, updates the layers' potentials and outputs one final prediction.}
\label{fig_arch2d}
\end{figure*}

We used four convolutional networks, noted A, B, C and D. We compared 3D convolutional neural networks with 2D spiking neural networks, with and without sparse operations. The architectures are detailed in Table \ref{tab:architectures} and the architecture of our SNN for a single timestep is illustrated in Fig.~\ref{fig_arch2d}.

\begin{table}[tb]
\renewcommand{\arraystretch}{1.3}
\centering
\caption{Architectures of the four models studied}
\begin{tabular}{l|lllll}
\hline
\textbf{Model} & \textbf{Layer 1} & \textbf{Layer 2} & \textbf{Layer 3} & \textbf{Layer 4}\\ \hline
A. CNN & \texttt{4c3-bn} & \texttt{8c3-bn} & \texttt{8c3-bn} &\texttt{16c3-bn} \\
B. Sparse CNN & \texttt{4sc3-bn} & \texttt{8sc3-bn}& \texttt{8sc3-bn} &\texttt{16sc3-bn} \\ 
C. SNN & \texttt{4c5} & \texttt{8c5} & \texttt{8c3} & \texttt{16c3-do} \\ 
D. Sparse SNN & \texttt{4sc5} & \texttt{8sc5} & \texttt{8sc3} & \texttt{16sc3-do} \\ \hline
\end{tabular}
\label{tab:architectures}
\end{table}

The CNNs use 3D kernels of size $3\times3\times3$ while the SNNs use 2D kernels of size $k\times k$ with $k=5$ for the first two convolutions and $k=3$ for the last two. Every convolutional layer has a stride of $2\times2$ on the spatial dimensions. The time dimension is not strided and kept constant with padding for the CNNs. The notation \texttt{sc} denotes the use of a sparse convolution, \texttt{bn} denotes the use of a batch normalization layer and \texttt{do} denotes the use of a dropout layer with 0.5 probability. Since we are training the parameter $b$, all spiking convolutional layers do not use biases. The activation function for the CNNs is ReLU and for the SNNs it is the Heaviside step function. Each network has a final fully-connected layer for classification.

\subsection{Implementation Details}

\subsubsection{Training}
All models were trained using the Rectified Adam optimizer \cite{radam} with a $1e^{-5}$ weight decay and an initial learning rate of $5e^{-3}$ for the models A, B, C, $1e^{-2}$ for the model D. The learning rate decayed by a factor $0.7$ every 2 epochs for models A, B, C. For model D, a cosine annealing scheduler with warm restarts every 30 epochs was used. The batch size was fixed to 16 for models A, B, C and to 48 for model D. Following \cite{s2net}, the leak and threshold parameters $\beta$ and $b$ of each layer were initialized to $0.7$ and $0.3$. Values of $b$ greater than $0.3$ prevent the learning as the layers become completely silent (i.e. emit zero spikes). Scale parameter $\alpha$ was fixed to $3$ after grid search. At each iteration, we clipped $\beta$ to $[0,1]$ and $b$ to $[0,+\infty[$. Norm of the gradient values were clipped to a maximum of 5. Presented results represent the best ones among 10 runs.

\subsubsection{Hardware}
All trainings were done using a 8-GB Tesla M60 and a 48-core Intel Xeon Gold 6126.

\subsection{Results\label{results}}

The results of the training of our 4 models are presented in Table \ref{tab:dvs128results}. The sparse SNN achieve the best test accuracy, beating the CNNs by 1.5\%. It is interesting to note that CNNs with 3D kernels have obtained competitive results of around 90\% test accuracy, in only 10 epochs of training. CNNs overfitted very rapidly, in spite of the high learning rate and weight decay, thus the batch normalization layers have been essential for proper learning.

The SNN has approximately 2.5 times longer training time per epoch than the CNN, due to its sequential processing. While the CNN process the whole sequence in one go, the SNN has to process each timestep sequentially. The best accuracies were achieved by the SNNs, demonstrating that taking advantage of the temporal nature of the data leads to better results.

The sparse CNN and SNN both achieves better test accuracy than their dense counterparts. For the CNNs, the training time is even divided by 2, showing a great benefit of switching from dense operations to sparse operations. For the SNNs, training one epoch is faster for the sparse SNN, but the training requires more epochs leading to an overall longer training.

To validate our timestep-wise approach, we trained an additional SNN with the same architecture and hyperparameters as our model C but using the layer-wise approach used in \cite{s2net}. Our SNN accuracy was better and the training was around 4 times faster per epoch, confirming the division of the training time by the number of convolutional layers.

\begin{table}[tb]
\renewcommand{\arraystretch}{1.3}
\centering
\caption{Test accuracy, epochs, and training time}
\begin{tabular}{l|ccc}
\hline
\textbf{Model} & \textbf{Accuracy} & \textbf{Epochs} & \textbf{Training time (per epoch)} \\ \hline
A. CNN & 90.28\% & 10 & 376s (38s) \\ 
B. Sparse CNN & 90.63\% & 10 & \textbf{228s (23s)} \\ 
C. SNN & 90.28\% & 12 & 1232s (95s) \\ 
D. Sparse SNN & \textbf{92.01\%} & 31 & 2639s (85s) \\  \hline
SNN \cite{s2net} & 87.50\% & 18 & 6332s (345s) \\ \hline
\end{tabular}
\label{tab:dvs128results}
\end{table}

\begin{table*}
\renewcommand{\arraystretch}{1.3}
\centering
\caption{Averaged non-zero activations (sparsity) after each layer during inference on the test set, over 150 timesteps}
\begin{tabular}{l|cc|cc|cc|cc|c}
\hline
\textbf{Model} & \multicolumn{2}{c}{\textbf{Conv 1}} & \multicolumn{2}{c}{\textbf{Conv 2}} & \multicolumn{2}{c}{\textbf{Conv 3}} & \multicolumn{2}{c}{\textbf{Conv 4}} & \textbf{Total} \\ \hline
A. CNN & 643k & (26\%) & 357k & (29\%) & 108k & (35\%) & 49k & (31\%) & 1,157k \\ 
B. Sparse CNN & 58k & (2.4\%) & 83k & (7\%) & 40k & (13\%) & 38k & (25\%) & 219k \\ 
C. SNN & \textbf{28k} & \textbf{(1.1\%)} & 26k & (2.1\%) & 14k & (4.6\%) & 13k & (8.5\%) & 81k \\ 
D. Sparse SNN & 53k & (2.1\%) & \textbf{7.2k} & \textbf{(0.6\%)} & \textbf{5.5k} & \textbf{(1.8\%)} & \textbf{1.7k} & \textbf{(1.1\%)} & \textbf{67.4k} \\ \hline
SNN \cite{s2net} & 38k & (1.5\%) & 61k & (4.9\%) & 17k & (5.5\%) & 16k & (10.4\%) & 132k \\ \hline
\end{tabular}
\label{tab:sparsity}
\end{table*}

We also measured the sparsity (i.e. the non-zero activations) during inference over 150 timesteps in each of the 4 convolutional layers, for each model. The results are presented in Table \ref{tab:sparsity}. 

Even with a convenient ReLU activation function that generates less non-zero activations than other activation functions, the CNN has a sparsity of around 30\% across the network, leading to nearly 1,157 non-zero activations. For a 99\% sparse input data, this result shows that traditional convolutional layers damage the input sparsity, making the use of CNNs to process event data unattractive.

Using sparse convolutions inside the CNN results in a significant improvement in the network sparsity. The first layer is 10 times sparser, better preserving the sparsity of the event data. The following layers also show an improvement, resulting in a total of 219k non-zero activations, more than 5 times sparser than the classical CNN. This result alone justifies the advantage of using sparse convolutional networks for the processing of event data.

The SNNs achieve impressive sparsity: a maximum of 8.5\% of the neurons spike at each layer, a number that even goes down to 0.6\% for some layers. This results in a total number of spikes smaller than 81k, 3 times sparser than the sparse CNN and 15 times sparser than the CNN. The sparse SNN generates an even sparser network, with a reduction of 13k spikes compared to the dense SNN. Both models generate twice as sparse networks as a SNN using the algorithm presented in \cite{s2net}.

To the best of our knowledge, our method represents the state-of-the-art in matter of sparsity for the DVS128 Gesture dataset.\\

\begin{table*}
\renewcommand{\arraystretch}{1.3}
\centering
\caption{Comparison with other SNNs}
\begin{tabular}{l|l|ccccc}
\hline
\textbf{Model} & \textbf{Network} & \textbf{\# parameters} &  \textbf{Training iterations} & \textbf{Accuracy} & \textbf{Real-time data} & \textbf{Simple operations}  \\ \hline
SLAYER \cite{slayer}  & 8 layers & Unknown & 270k & 93.64\%  &  \checkmark  & $\times$         \\  
SCRNN \cite{scrnn}    & \texttt{32c5-64c3-128c3-1024-512-11} & \textgreater{662k} & 100 epochs & 92.01\% &  \checkmark   & $\times  $          \\ 
DECOLLE \cite{decolle} & \texttt{64c7-128c7-128c7-11}  & \textgreater{51k} & 160k & 95.54\% & \checkmark  &   \checkmark \\ 
PLIF SNN \cite{fangtim} & \texttt{5*\{128c3\}-512-110} & \textgreater{1,110k} & 74k & 97.57\% & $\times$ & \checkmark \\
This work  & \texttt{4sc5-8sc5-8sc3-16sc3-11} & 14k & 1.3k & 92.01\% & \checkmark & \checkmark    \\ \hline
\end{tabular}
\label{tab:final_comparison}
\end{table*}

Table \ref{tab:final_comparison} compares our Sparse SNN with SOTA SNNs from the literature. When the number of parameters was unspecified, we assumed that the final output feature map was of size $4 \times 4$, probably underestimating the real number of parameters. Although our model is not better in terms of accuracy, it is still competitive with up to 100 times less parameters and training iterations. We reviewed if models are able to use event data coming directly coming from event camera  with minimal preprocessing, representing a real time situation. Reference \cite{fangtim} gathers frames of events depending on the number of events, which is not compatible with a real time situation where the number of events can vary greatly over time (and prevent a prediction if the number is not sufficient). Even if others SNNs are capable of using real-time event data, our model is the only one using binary event data. We also reviewed if the operations used in the models are compatible with a low-power neuromorphic implementation, i.e. if the operations are simple. Since our work only uses convolutions, multiplication and addition, it remains easily the simplest network to implement on neuromorphic hardware. Other works include the use of mean/max pooling, special memory cells... We could not compare sparsity with other SNNs since this work is the first providing such a measurement.

\subsection{Impact of strided convolutions\label{stride}}

Our choice of using convolutions with a stride of 2 rather than a pooling operation is mainly motivated by the absence of this operation on the targeted hardware \cite{nassim2}, \cite{thesenassim}. Using a stride of 2 results in half the number of operations performed during convolutions, which is obviously an interesting property for an embedded implementation. We also had the intuition that using convolutions with a stride of 2 would generate a sparser network. 

To validate our choice, we replaced the strided convolutions by convolutions with a stride of 1 followed by a $2\times2$ max pooling layer in our model D, the sparse SNN. We measured the maximum accuracy and the sparsity of this network over 5 runs, and compared it with our reference model. The results are presented in Table \ref{tab:stride}.

Using pooling layers results in a loss of accuracy of 1\%, and more importantly in twice as many spikes generated across the network compared to the use of strided convolutions. The use of strided sparse convolutions on event data is therefore of interest to obtain a better accuracy, a reduced number of operations and a sparser network.

\begin{table}[tb]
\renewcommand{\arraystretch}{1.3}
\centering
\caption{Comparison between strided convolutions \\ and pooling for sparse SNNs}
\begin{tabular}{l|cc}
\hline
\textbf{Model}                   & \textbf{Accuracy} & \textbf{\# spikes}\\ \hline
Sparse SNN with stride  & \textbf{92.01}\%  & \textbf{67.4k}\\
Sparse SNN with pooling & 90.97\%  & 156.4k\\ \hline
\end{tabular}
\label{tab:stride}
\end{table}

\section{Discussion on real-time inference on neuromorphic hardware}

Previous works have showed that implementing spiking neural networks on specialized hardware can brings a 50\% gain in energy consumption over traditional neural networks, for the same accuracy \cite{lyesijcnn}. These gains could be further increased with the use of neuromorphic hardware such as Intel Loihi \cite{loihi}. This makes their embedded use particularly interesting in power-limited environment such as cars. Algorithms embedded in a car have high constraints in terms of accuracy, latency and energy consumption. A fully event-based approach with data coming from event cameras, processed by spiking neural networks running on neuromorphic hardware represents an ideal solution. 

Because of its timestep by timestep functioning and the operations used, our method represents a plausible implementation of such an event-based approach. Indeed, spiking convolutions have successfully been implemented on low-power hardware \cite{nassim2}, \cite{thesenassim}, and our model is able to process a continuous stream of events directly coming from an event camera. As mentioned previously, our method is able to output a final prediction after any defined timestep, representing a realistic real-time inference. 

\begin{figure}[tb]
\centerline{\includegraphics[scale=0.34]{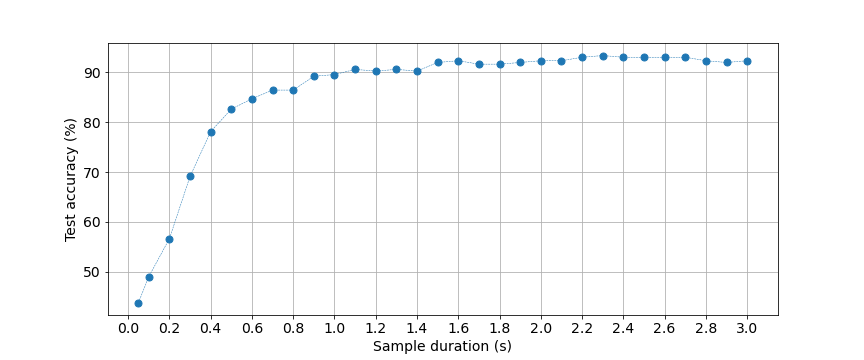}}
\caption{Test accuracy for samples of different durations}
\label{fig:ts}
\end{figure}

We validate this claim by computing the test accuracy on DVS128 Gesture of our sparse SNN with different number of timesteps, ranging from 5 to 300. We are thus doing inference on samples lasting from 50ms to 3.0s. Results are presented in Fig.~\ref{fig:ts}.

The accuracy of our model is quite robust to the duration of the samples, as it achieves a 86.46\% test accuracy for 0.7s samples and even 82.64\% for 0.5s samples. Accuracy drops rapidly when the number of timesteps is small (inferior to 50). Increasing the sample duration leads almost always to a better accuracy until the 1.5s mark, then the accuracy stays relatively constant. Our model even achieves 93.40\% test accuracy for 2.3s samples. Therefore, it benefits from using more timesteps at inference than during training. Furthermore, as the number of timesteps is directly related to the number of operations performed, our work allows a trade-off between accuracy, power consumption, latency and memory.

This behavior is noteworthy for an embedded real-time inference: intermediate results can be obtained at the same rate as data is captured, and prediction can either be continued or be stopped according to the confidence, enabling even more energy savings without hindering the accuracy.

\section{Conclusion and future works}

We presented a timestep-wise approach to build sparse spiking neural networks learning directly from binary event data. Our method generates highly sparse networks and is able to output a prediction at any timestep, two essential characteristics for a real-time inference on neuromorphic low power hardware. We validated our approach on the neuromorphic DVS128 Gesture dataset, achieving 93.40\% test accuracy with a sparse spiking convolutional network, where the whole network does not generate more than 450 spikes per timestep. 

Future works would include the real implementation of inference on a low-power neuromorphic device \cite{loihi}, \cite{nassim}, processing on the fly continuous data coming from an event sensor. The reduction in training time provided by our method also enables the learning of SNNs on automotive event datasets \cite{chen}, broadening the use-cases of this type of network for embedded real-time applications.




\section*{Acknowledgment}

This material is based upon work supported by the French
technological research agency (ANRT) through a CIFRE thesis in collaboration with Renault.


\begin{thebibliography}{00}

\bibitem{neftci} E. Neftci, H. Mostafa, and F. Zenke, “Surrogate gradient learning in spiking neural networks: bringing the power of gradient-based optimization to spiking neural networks,” IEEE Signal Processing Magazine, vol. 36, no. 6, pp. 51–63, 2019.

\bibitem{s2net} T. Pellegrini, R. Zimmer, and T. Masquelier, ``Low-activity supervised convolutional spiking neural networks applied to speech commands recognition,`` IEEE Spoken Language Technology Workshop, 2021, in press.

\bibitem{dvs128} A. Amir et al., "A low power, fully event-based gesture recognition system," IEEE Conference on Computer Vision and Pattern Recognition, 2017.

\bibitem{steering2018} A. I. Maqueda, A. Loquercio, G. Gallego, N. Garcia, and D. Scaramuzza,``Event-based vision meets deep learning on steering prediction for self-driving cars,`` IEEE Conference on Computer Vision and Pattern Recognition, 2018.

\bibitem{hots} X. Lagorce, G. Orchard, F. Galluppi, B. E. Shi, and R. B. Benosman, "HOTS: a hierarchy of event-based time-surfaces for pattern recognition," IEEE Transactions on Pattern Analysis and Machine Intelligence, vol. 39, no. 7, pp. 1346-1359, July 2017.

\bibitem{iwe} G. Gallego, H. Rebecq, and D. Scaramuzza, “A unifying contrast maximization framework for event cameras, with applications to motion, depth, and optical flow estimation,” IEEE Conference on Computer Vision and Pattern Recognition, 2018.

\bibitem{reconstruction} H. Rebecq, R. Ranftl, V. Koltun, and D. Scaramuzza, “High speed and high dynamic range video with an event camera,” Transactions on Pattern Analysis and Machine Intelligence, 2019.

\bibitem{voxel} P. Bardow, A. J. Davison, and S. Leutenegger, “Simultaneous optical flow and intensity estimation from an event camera,” IEEE Conference on Computer Vision and Pattern Recognition, 2016.

\bibitem{slayer} S. B. Shrestha, and G. Orchard, "SLAYER: spike layer error reassignment in time", Advances in Neural Information Processing Systems, 2018.

\bibitem{scrnn} Y. Xing, G. Di Caterina, J. Soraghan, "A new spiking convolutional recurrent neural network (SCRNN) with applications to event-based hand gesture recognition", Frontiers in Neuroscience, vol. 14, pp. 1143, 2020.

\bibitem{decolle} J. Kaiser, H. Mostafa, and E. Neftci, "Synaptic plasticity dynamics for deep continuous local learning (DECOLLE)," Frontiers in Neuroscience, May 2020.

\bibitem{fangtim} W. Fang, Z. Yu, Y. Chen, T. Masquelier, T. Huang, and Y. Tian, "Incorporating learnable membrane time constant to enhance learning of spiking neural networks", arXiv:2007.05785, 2020, unpublished. 

\bibitem{snn} W. Maass, "Networks of spiking neurons: the third generation of neural network models", Neural Networks, vol. 10, no. 9, pp. 1659-1671, 1997.

\bibitem{lif} A. Delorme, J. Gautrais, R. Van Rullen, and S. Thorpe, “Spikenet: A simulator for modeling large networks of integrate and fire neurons,” Neurocomputing, vol. 26, pp. 989–996, 1999.

\bibitem{hh} A. L. Hodgkin and A. F. Huxley, “A quantitative description of membrane current and its application to conduction and excitation in nerve,” The Journal of physiology, vol. 117, no. 4, pp. 500–544, 1952.

\bibitem{kherad} S.R. Kheradpisheh M. Ganjtabesh, S.J. Thorpe, and T. Masquelier, "STDP-based spiking deep convolutional neural networks for object recognition," Neural Networks, 2017.

\bibitem{spikeprop} S. M. Bohte, J. N. Kok, and H. La Poutre, “Error backpropagation in temporally encoded networks of spiking neurons,” Neurocomputing, vol. 48, no. 1, pp. 17–37, 2002.

\bibitem{sparseCNN} B. Liu, M. Wang, H. Foroosh, M. Tappen and M. Penksy, "Sparse convolutional neural networks," IEEE Conference on Computer Vision and Pattern Recognition, pp. 806-814, 2015.

\bibitem{sparseconvnet} B. Graham, M. Engelcke, and L. van der Maaten, "3D semantic segmentation with submanifold sparse convolutional networks," IEEE Conference on Computer Vision and Pattern Recognition, 2018.

\bibitem{me} C. Choy, J. Gwak, and S. Savarese, "4D spatio-temporal convnets: Minkowski convolutional neural networks," IEEE Conference on Computer Vision and Pattern Recognition, pp. 3075-3084, 2019.

\bibitem{eventscnn} N. Messikommer, D. Gehrig, A. Loquercio, and D. Scaramuzza, "Event-based asynchronous sparse convolutional networks," European Conference on Computer Vision, 2020.

\bibitem{dvs128camera} P. Lichtsteiner, C. Posch, and T. Delbruck, ``A 128 X 128 120db 30mw asynchronous vision sensor that responds to relative intensity change,`` IEEE International Solid State Circuits Conference - Digest of Technical Papers, pp. 2060-2069, February 2006.

\bibitem{prophesee1Megapixel} E. Perot, P. de Tournemire, D. Nitti, J. Masci, and Amos Sironi, "Learning to detect objects with a 1 Megapixel event camera," Advances in Neural Information Processing Systems, 2020.

\bibitem{pool1} X. Rong, et al. "An Event-Driven Categorization Model for AER Image Sensors Using Multispike Encoding and Learning," IEEE Transactions on Neural Networks and Learning Systems, vol. 31, no. 9, pp. 3649-3657, 2020.

\bibitem{pool2} B. Zhao, R. Ding, S. Chen, B. Linares-Barranco, H. Tang, "Feedforward Categorization on AER Motion Events Using Cortex-like Features in a Spiking Neural Network," IEEE Transactions on Neural Networks and Learning Systems, vol. 26, no. 9, pp. 1963-1978, 2014.

\bibitem{radam} L. Liu, H. Jiang, P. He, W. Chen, X. Liu, J. Gao, and J. Han, ``On the variance of the adaptive learning rate and beyond,`` International Conference on Learning Representations, 2020.


\bibitem{lyesijcnn} L. Khacef, N. Abderrahmane, and B. Miramond, "Confronting machine-learning with neuroscience for neuromorphic architectures design", International Joint Conference on Neural Networks, pp. 1-8, 2018.


\bibitem{nassim2}  N. Abderrahmane and B. Miramond, "Neural coding: adapting spike generation for embedded hardware classification," International Joint Conference on Neural Networks (IJCNN), Glasgow, United Kingdom, 2020.

\bibitem{thesenassim} N. Abderrahmane, "Hardware design of spiking neural networks for energy efficient brain-inspired computing," 2020.

\bibitem{loihi} M. Davies et al., "Loihi: a neuromorphic manycore processor with on-chip learning," in IEEE Micro, vol. 38, no. 1, pp. 82-99, January/February 2018.

\bibitem{nassim} N. Abderrahmane, E. Lemaire, and B. Miramond, "Design space exploration of hardware spiking neurons for embedded artificial intelligence", Neural Networks, vol. 121, pp. 366-386, 2020.

\bibitem{chen} G. Chen, H. Cao, J. Conradt, H. Tang, F. Rohrbein and A. Knoll, "Event-based neuromorphic vision for autonomous driving: a paradigm shift for bio-inspired visual sensing and perception," in IEEE Signal Processing Magazine, vol. 37, no. 4, pp. 34-49, July 2020.



\end{thebibliography}
\end{document}